\documentclass[a4paper, 10pt, conference]{ieeeconf}      

\IEEEoverridecommandlockouts                              

\usepackage{epsfig} 
\usepackage{amsmath} 
\usepackage{amssymb}  
\usepackage{subcaption}
\usepackage{tabularx}
\usepackage{graphicx}
\usepackage{ragged2e}
\usepackage{xcolor}

\title{Design of a Multi-Degree-of-Freedom Elastic Neck Exoskeleton for Persons with Dropped Head Syndrome}

\author{Santiago Price Torrendell$^{1}$, Yang Chen$^{1}$, Hideki Kadone$^{2}$, Modar Hassan$^{3}$, and Kenji Suzuki$^{3}$
\thanks{$^{1}$Santiago Price Torrendell and Yang Chen are with the Graduate School of Systems \& Information Engineering,
        University of Tsukuba, Ibaraki, Japan
        {\tt\small santiago@ai.iit.tsukuba.ac.jp} , {\tt\small chenyang@ai.iit.tsukuba.ac.jp}}%
\thanks{$^{2}$Hideki Kadone is with the Center for Innovative Medicine and Engineering, University of Tsukuba Hospital, Japan
        {\tt\small kadone@ccr.tsukuba.ac.jp}}%
\thanks{$^{3}$Modar Hassan and Kenji Suzuki are with the Faculty of Systems, Information and Engineering, University of Tsukuba, Tsukuba, Japan
        {\tt\small modar@iit.tsukuba.ac.jp}, {\tt\small kenji@ieee.org}}%
}

\begin{document}

\maketitle
\thispagestyle{empty}
\pagestyle{empty}

\begin{abstract}
Nonsurgical treatment of Dropped Head Syndrome (DHS) incurs the use of collar-type orthoses that immobilize the neck and cause discomfort and sores under the chin. Articulated orthoses have the potential to support the head posture while allowing partial mobility of the neck and reduced discomfort and sores. This work presents the design, modeling, development, and characterization of a novel multi-degree-of-freedom elastic mechanism designed for neck support. This new type of elastic mechanism allows the bending of the head in the sagittal and coronal planes, and head rotations in the transverse plane. From these articulate movements, the mechanism generates moments that restore the head and neck to the upright posture, thus compensating for the muscle weakness caused by DHS. The experimental results show adherence to the empirical characterization of the elastic mechanism under flexion to the model-based calculations. A neck support orthosis prototype based on the proposed mechanism is presented, which enables the three before-mentioned head motions of a healthy participant, according to the results of preliminary tests.

\end{abstract}

\section{INTRODUCTION} 
Dropped Head Syndrome (DHS) is a set of conditions caused by a wide range of muscular, neurological, or neuro-muscular disorders \cite{Alhammad2020}. These dysfunctions tend to induce a severe kyphotic deformity of the cervicothoracic spine, which hinders or disables the neck extension of the affected persons \cite{Kadone2020}. This results in considerable restrictions on ambulation, activities of daily living (ADL), and social interactions \cite{Martin2011}. In some cases, this deformation is corrected by providing mechanical support to hold the head, which is frequently done by either surgical procedures or passive orthosis. Both options restrict partially or totally the mobility of the user’s neck, severely impacting their quality of life (QoL)  \cite{Kadone2020}. There is a clear clinical need for a device that fixes the user’s head posture and enables or assists neck mobility safely, beyond the current solutions with their evident limitations.

Neck motion plays an essential role in several activities of daily living. In an observational study,  Cobian et al. \cite{Cobian2013} points out that a healthy person bends the head an average of 58 degrees on the sagittal plane, 45 degrees on the coronal plane, and rotates it 70 degrees in both directions on the transverse plane during daily activities. Consequently, the neck movement restriction caused by muscle weakness in patients with DHS severely affects their QoL, an impairment that traditional treatments cannot completely mitigate. The first line of management involves the prescription of a neck brace that fixes the head of the patient in an upright posture. This has a negative impact on the patient's comfort and contributes to muscle weakness \cite{Jansen2013}, which may aggravate the kyphotic deformity. If the patient does not recover after the neck immobilization and neurological diseases have been ruled out, the patient’s vertebrae are surgically fixed to keep their head in a straight position. Besides the movement restriction, the inherited risk of surgery becomes even higher in older people which is the age group mainly affected by DHS \cite{Brodell2020}, because of their poorer structure which requires more implants to ensure the structure’s mechanical stability \cite{Martin2011}.       

\begin{figure}[t]
    \begin{center}
    \includegraphics[width=0.9\linewidth]{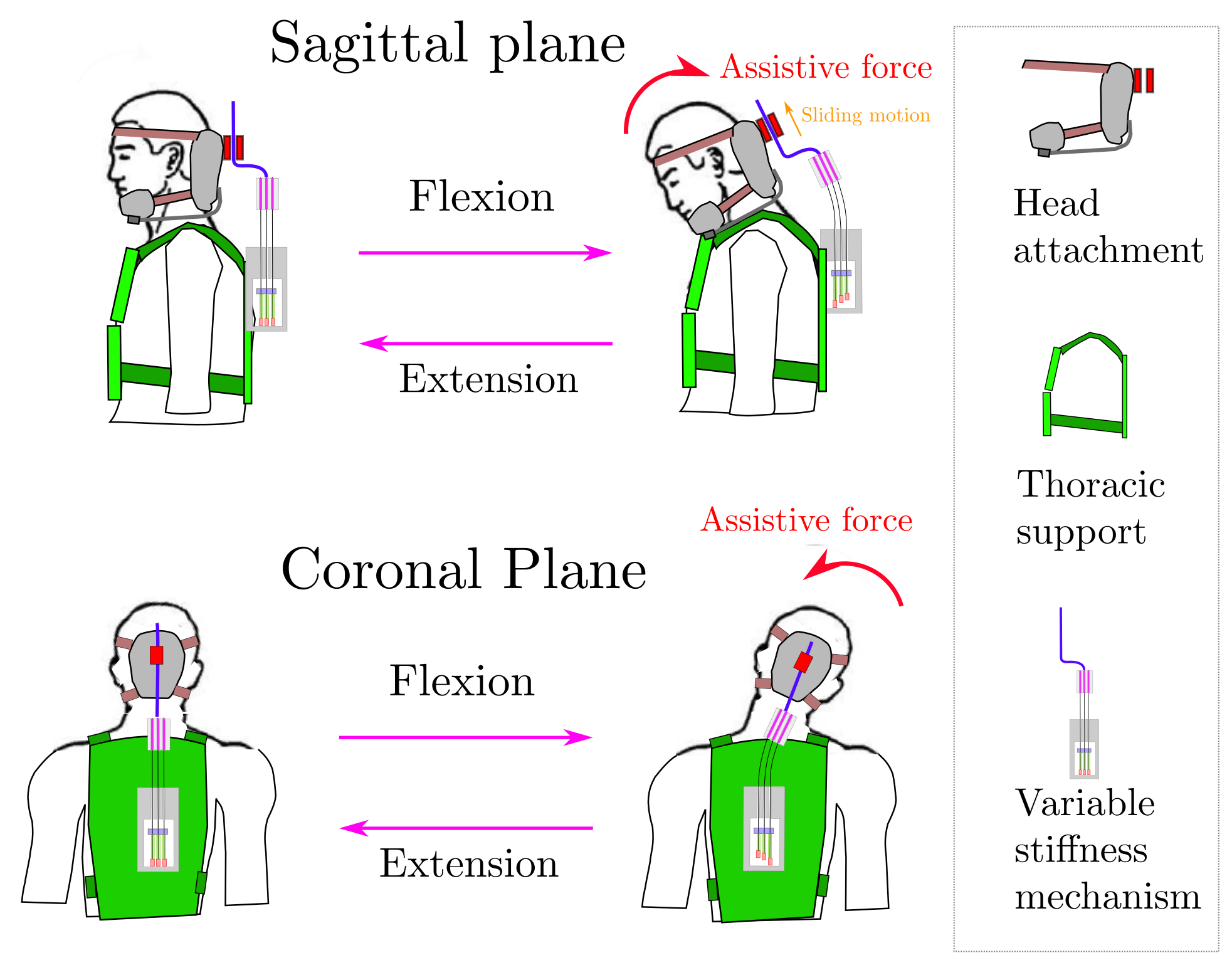} 
    \caption{Illustration of neck mobility with the proposed elastic mechanism and components of the exoskeleton device. The assistive force is regulated by adjusting the stiffness of the elastic mechanism.}
    \label{fig:head_assitive_eff}
    \end{center}      
    \vspace{-0.6cm}
\end{figure} 

In contrast with the treatment methodology explained above, several researchers present different devices for treating DHS caused by muscular or neuro-muscular issues. Kadone et al. \cite{Kadone2020} proposed the use of a robotic lower limbs exoskeleton as an alternative conservative treatment for these patients. Their research showed that gait training sessions assisted by the exoskeleton robot helped to reduce the severity of DHS. Other works proposed assisting the neck motion with articulated or compliant neck braces (\cite{Mahmood2021},\cite{Zhang2017},\cite{Baxter2016}). Mohammad et al. \cite{Mahmood2021} developed a neck support orthosis with a gravity support mechanism, their results showed a significant reduction in the upper trapezius muscle activity. Haohan et al. \cite{Zhang2017} developed a neck orthosis with multiple parallel mechanical linkage chains to stabilize the head posture relative to the shoulders. The purpose of the design is to measure the head posture and movement, and allowed 70\% of the range of head rotations. The authors explored several actuation strategies for such a device considering the use of electric motors controlled by different interfaces (\cite{Zhang2018},\cite{Zhang20182}) as well as the use of passive torsion springs \cite{Zhang20183}. The motorized version of the device has been evaluated in DHS patients for studying the relation between the neck muscle activation pattern with the head motion \cite{Zhang2019}, and their ability to regain neck-head control, showing positive results \cite{Zhang2022}.
The Snood Sheffield Support is a snood-shaped neck orthosis made of fabric whose stiffness can be changed with removable structures enabling support customization of the neck support to each user case \cite{Baxter2016}. Although the design is not centered on supporting the head in different positions the device enables the extension, flexion, and axial rotation of the neck \cite{Pancani2016}.
The devices mentioned above each step towards an articulated neck support orthosis to replace the fixed brace that completely immobilizes the neck of DHS patients. However, they do not cover all the requirements of degrees of freedom, range of motion, and portability for the support of DHS. In this context, we propose that a compliant passive mechanism for the brace is a promising approach to benefit DHS. Recent works used flexible beams in a lower back support exoskeleton which showed positive results in assisting lifting objects from the ground (\cite{Nf2018}, \cite{chang2020}). In this work, we propose a multi-degree-of-freedom (multi-dof) elastic mechanism capable of providing restorative moments in multiple anatomical directions, which has the potential to support head stability while allowing a reasonable range of neck mobility. The following sections present the design, modeling, development, and characterization of the elastic mechanism, the development of a neck support orthosis prototype that incorporates the elastic mechanism, and preliminary experiments on healthy participants.

\section{methods}  
This work investigated the concept of a multi-dof elastic mechanism because it has the potential to assist the head while minimizing movement restrictions. The mechanism deploys a hyper-redundant structure that continuously bends to conform with the neck posture, which makes it unobtrusive to human back anatomy and has the potential to overcome the limitations in terms of ergonomics and range of motion of previous devices \cite{Yang2019}. This elastic mechanism does not require gearing or power transmission which makes the system lighter and smaller than traditional mechanisms, features that are highly relevant for this application \cite{Steenbergen2011}.


\subsection{Mechanism design}
\subsubsection{Actuation principle}
The main function of the device is to provide assistive forces that compensate for the user's neck muscle weakness while maintaining neck mobility. This assistive effect occurs when the user bends their head in the sagittal or coronal plane, as shown in Fig. \ref{fig:head_assitive_eff}, or in any intermediate condition.
During flexion, the elastic mechanism bends following the cervical spine's curve, producing restoring moments transmitted to the head via belt attachments. The mechanism includes elastic bars attached to the head and upper back which combined with a set of compression springs and a preloading mechanism regulates the system's total stiffness. The elastic mechanism and head attachment are connected by a free prismatic joint. The sliding system minimizes spine compressing force by aligning with the skull's vertical axis, preventing forces transmission along its longitudinal axis.

\begin{figure}[t]
    \begin{center}
    \includegraphics[width=0.9\linewidth]{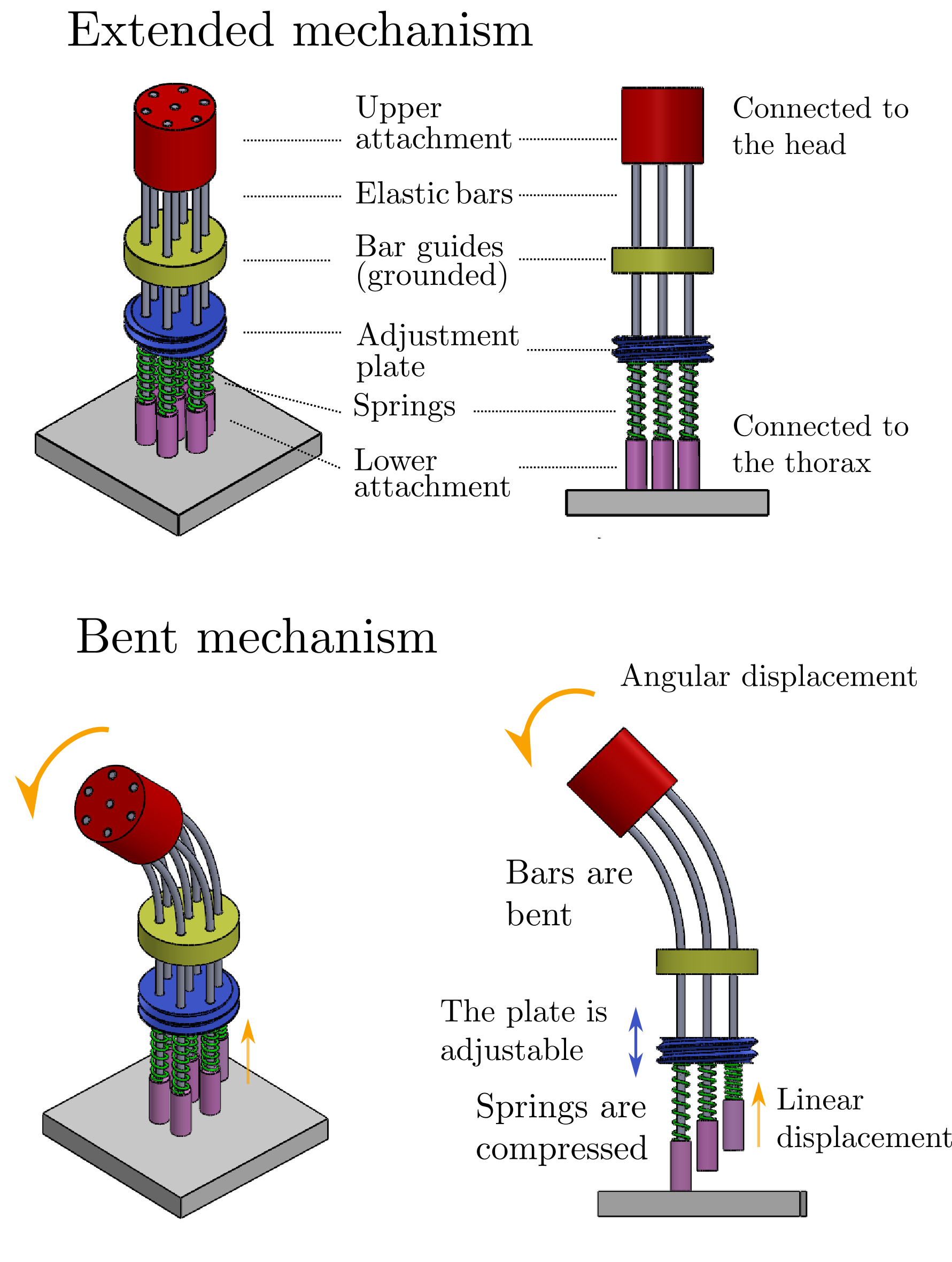} 
    \caption{Elastic mechanism behavior. The bars' deflection causes sliding of the lower end upwards which compresses the springs. Consequently, changing the preload of the springs by moving the adjustable plate regulates their compression forces, which varies the stiffness of the mechanism.}
    \label{fig:mech_scheme}
    \end{center} 
    \vspace{-0.7cm}
\end{figure} 

\subsubsection{Elastic mechanism based on flexible bars and compression springs}
The elastic mechanism compromises a set of elastic bars coupled to springs. Fig. \ref{fig:mech_scheme} shows a simplified model of the device where the upper attachment is linked with the head motion, the bar guide and the ground are attached to the thorax support and the bars connect both components. Each bar is rigidly attached to the upper attachment and goes through the bar guide, the adjustable plate, and the spring. The mechanism allows vertical movement of the bars which slide through the guides, compressing the spring between the lower attachment and the adjustable plate. The spring's preload is regulated by the mobile plate. When the system is bent, the bars deflect and translate vertically through the guides, compressing the springs against the plate. This response is analogous to a single beam under deflection, where each bar of the mechanism would represent a fiber of the single bar. Stephen Timoshenko explains in his book \cite{Tym1948} that in a rectangular beam under pure deflection the fibers at the concave side are compressed while on the convex side are stretched to make the global deformation compatible with the semi-arch shape observed in the experiments. In the mechanism, the system deflects similarly so the bars will be compressed or stressed to adapt to the arch shape. The lateral view of the bent system in Fig. \ref{fig:mech_scheme} shows that the springs at the concave side remain unchanged as the bar is compressed between the upper and lower attachment. To allow for the total deformation, the bars at the middle and convex side should extend which creates a stretching force between the upper attachment and the adjustable plate. In this condition, the compressed springs deform instead of bars due to their lower stiffness. The plate's position changes the springs' preload, altering the resisting force for bar displacement and controlling the system's compliance.

\begin{figure}[t]
    \begin{center}
    \includegraphics[width=0.9\linewidth]{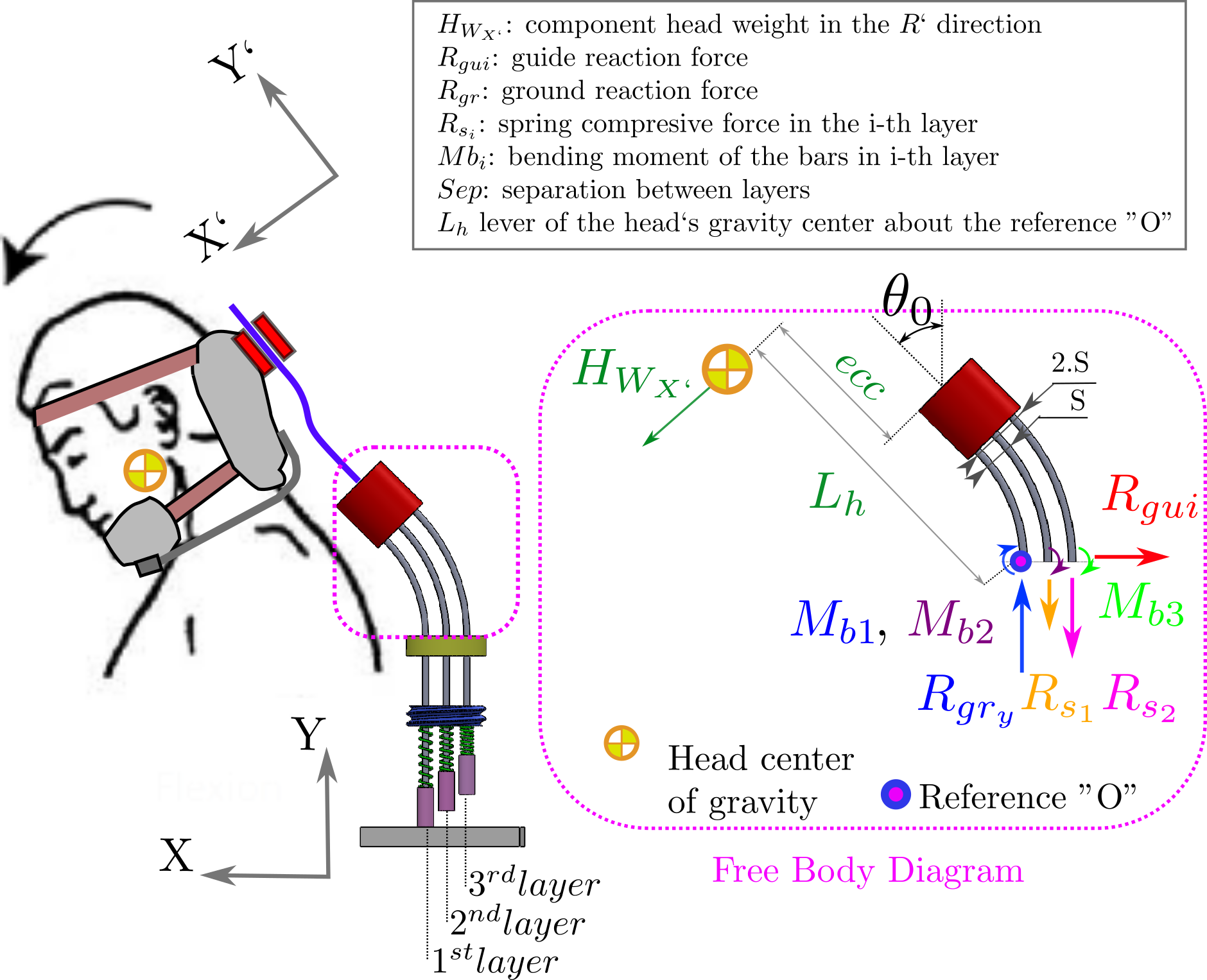} 
    \caption{Actuator modeling: the free body diagram of the system shows that the head weight component in the X' direction $H_{W_x}$, the bending moment of the bars $M_{b_i}$ and the springs compressing force $R_{s_i}$ are present when the system is bent to a given angle $\theta$. Evaluating the moment equilibrium about the reference "O" provides an expression that relates the $\theta$ with the assistive torque of the device.}
    \label{fig:act_FBD}
    \end{center} 
    \vspace{-0.6cm}
\end{figure} 

\subsection{Modeling}
The flexible bars and the springs govern the mechanical behavior of the elastic mechanism. Therefore we created a model of the device to elucidate how the mechanical properties of these two elements and their geometrical configuration affect the relationship between the bending angle and the provided assistive force. This relationship is presented in the free body diagram (FBD) shown in Fig. \ref{fig:act_FBD} where the forces and moments of the compliant elements counterbalance the assistive force acting on the head. It should be noted that this formulation does not consider the twisting of the system, a condition that is present during the axial rotation of the user's head. 

The assistive force provided by the elastic mechanism can be calculated through the moment's balance of the compliant elements in the FBD. By choosing the reference "O" for calculating the moments, the ground forces are excluded from the analysis. 

Getting the total sum of the moments gives the following equation: 
\begin{align}
\begin{split}
    \sum M_O
    &=0 \\
    &= M_{bars}+M_{springs}+M_{assist}\\
    &= 2\cdot M_{b_1}+ 3\cdot(M_{b_2}+R_{s_2}\cdot Sep)+2 \cdot (M_{b_3}\\& \; \; \;+F_{s_3}\cdot2\cdot Sep) + F_{assist}\cdot L_h
\end{split}
    \label{eq:mom_bal}
\end{align}
Where $M_{assist}$ is the assistive moment, and $F_{si}$ with $M_{b_i}$ are the spring reaction force and the bending moment of the bar in the i'th layer. The 1st layer includes the bars on the convex side, the 2nd layer includes the bars in the middle and the 3rd layer includes the bars on the concave side. The forces and moments depend on each element's deformation, which is geometrically linked with the elastic mechanism's deflection. The geometrical relation is established assuming the separation between the bars remains constant after the system is bent.

\begin{figure}[t]
    \begin{center}
    \includegraphics[width=0.9\linewidth]{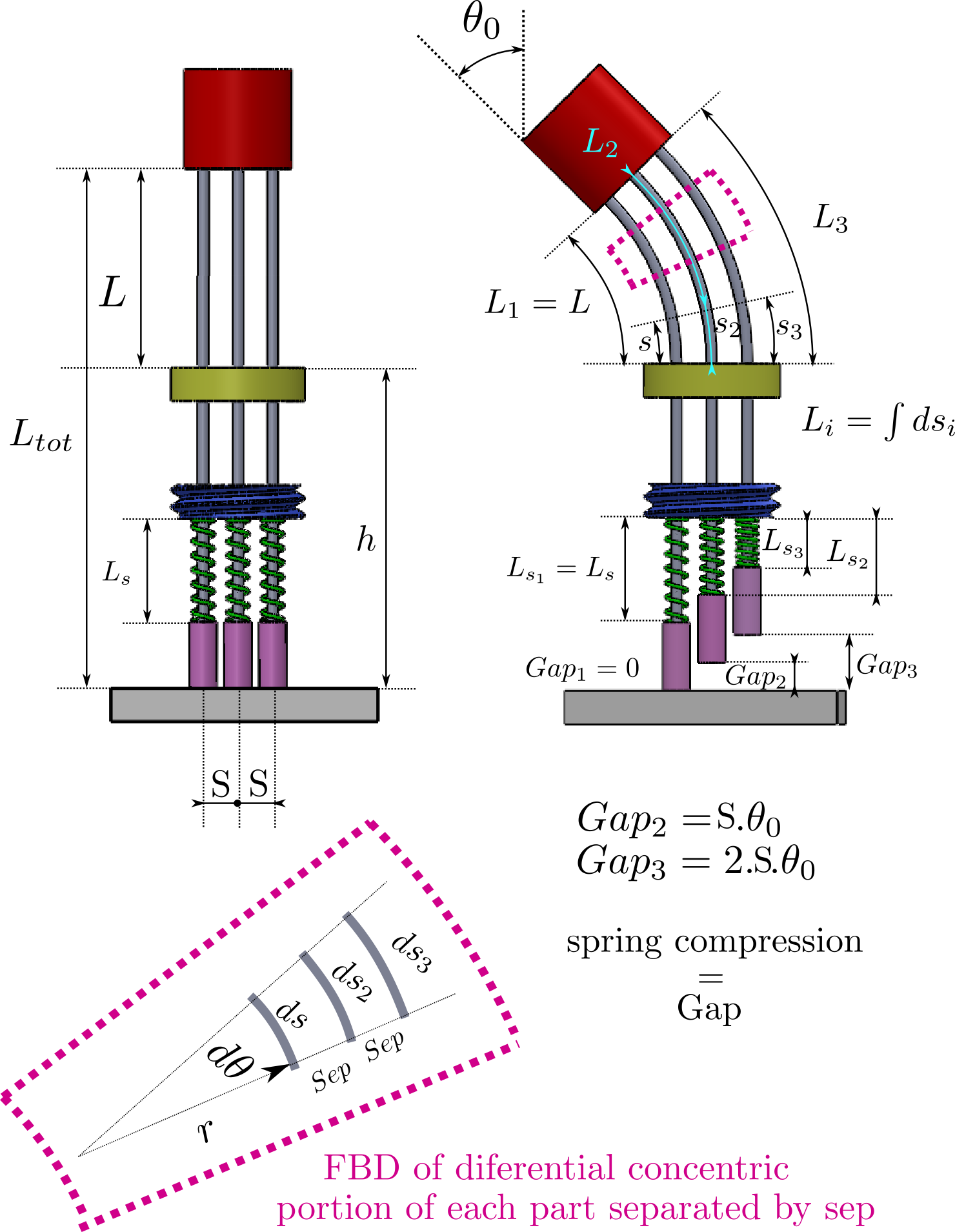} 
    \caption{Spring compression definition from the elastic mechanism dimensions in the extended and bent conditions. Under the hypothesis that the separation of the bars remains constant, it can be deduced that the spring compression $\delta L = L_{s}-L_{s_i}$ linearly depends on the bending angle $\theta$. }
    \label{fig:act_lengths}
    \end{center}   
    \vspace{-0.5cm}
\end{figure} 

The force provided by each spring is linear to its deformation, which is determined by the mechanism's geometry shown in Fig. \ref{fig:act_lengths}. From the scheme, it can be deduced that the compression is equal to the length change $L-L_i$ of each bar's bent portion. Thus, the spring deformation depends on $\theta$ as in Equation \ref{eq:spring_eq}:
\begin{align}
\begin{split}
F_{s_i}&=K \cdot (L_s - L_{s_i}) \\
F_{s_i}&=K \cdot (L_s - \int_0^{\theta_0}\dfrac{ds_i}{d\theta} \cdot d\theta)\\
F_{s_1}&=K \cdot (L_s - L_s) = 0\\
F_{s_1}&=K \cdot (L_s - \int_0^{\theta_0}(\dfrac{ds_1}{d\theta}+S)\cdot d\theta) = K \cdot S \cdot\theta_0\\
F_{s_2}&=K \cdot (L_s - \int_0^{\theta_0}(\dfrac{ds_1}{d\theta}+2\cdot S) \cdot d\theta) = K \cdot 2  S \cdot \theta_0\\
\end{split}
\label{eq:spring_eq}
\end{align}

Where $K$ is the spring constant, and $L_{s_i}$ is calculated by integrating the corresponding differential element in the diagram of Fig. \ref{fig:act_lengths}.

\begin{figure}[t]
    \begin{center}
    \includegraphics[width=0.9\linewidth]{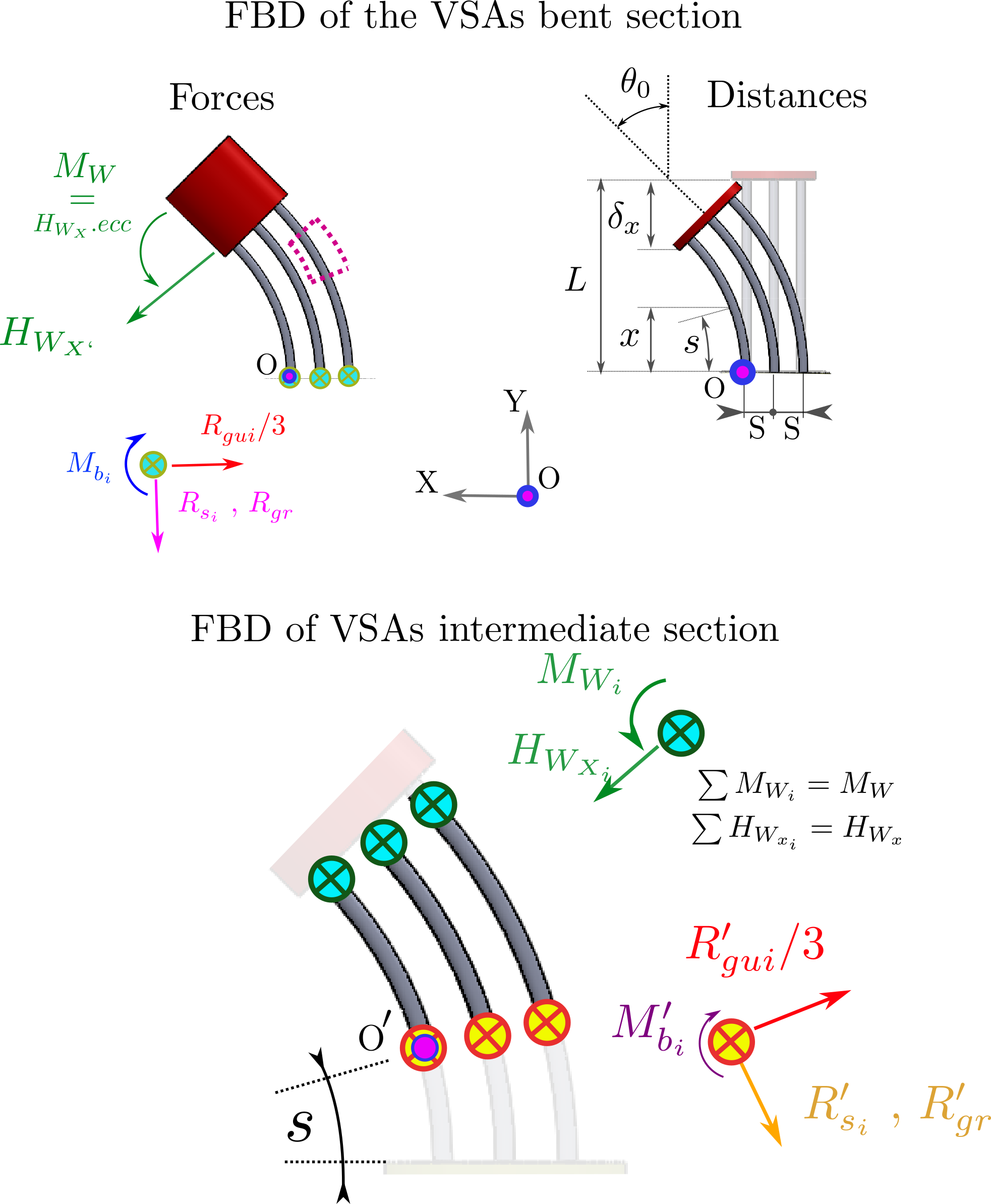} 
    \caption{Static analysis for linking the bars' bending moment with their mechanical properties and the elastic mechanism deflection angle. The FBDs in the first row shows the forces and geometrical parameters involved in the analysis. Below, the left FBD is used to get the relationship using simplifications elaborated in the diagram of the single bar. The physical cause of each force is mentioned in Fig. \ref{fig:act_FBD}}
    \label{fig:act_bending forces}
    \end{center}  
     \vspace{-5mm}
\end{figure}

The relationship between the bar's bending moment and its deflection is estimated by using the Bernoulli–Euler bending moment-curvature relationship:
\begin{equation}
    E.I\dfrac{d\theta}{ds} = M
    \label{eq:beam_eq}
\end{equation}
Where M and $\kappa = \dfrac{d\theta}{ds}$ are the bending moment and the curvature at any point of the beam, and I is the moment of inertia of the beam's cross-section about the neutral axis \cite{Belen2002}. The bending moment along the three bars counterbalances the elastic mechanism assistive force and the springs contribution shown in the FBD in Fig. \ref{fig:act_bending forces}. 
Evaluating the moment equilibrium of the system gives the differential equation that defines the bar's moment-curvature relationship.  \\
\begin{align}
    \begin{split}
    \sum M_0 &=-\dfrac{H_{W_x}\cdot  (L-X-\delta_x)}{cos(\theta_0)} - M_W  +\sum M_i+ M_{spr} \\ & = 0 \\ 
    & E\cdot I(2\cdot \dfrac{d\theta}{ds}+3\cdot  \dfrac{d\theta}{ds_2}+2\cdot  \dfrac{d\theta}{ds_3})= M_{bars} \\
    &\text{where} M_{bars} = \dfrac{H_{W_x}\cdot (L-X-\delta_x)}{cos(\theta_0)} + M_W - M_{spr} 
    \end{split}
\end{align}

Based on the equation of the bar total moment, it will be assumed that the bending moment along each bar is a fraction of $M_{bars}$ that does not depend on the arch length $s_i$. This allows us to decompose the problem and solve each bar relationship separately with Equation \ref{eq:beam_eq}. 

To solve the single bar problem it is decomposed into two sub-problems where the deflection of the bar is solved considering only the moment or the force separately, resulting in the following two equations: 

\begin{equation}
    \theta_i = \theta_{i_M} + \theta_{i_F} \\
    \label{eq:mom_eq_simp}
     \vspace{-4mm}
\end{equation} 

 \begin{equation}
     \text{where } E \cdot I\dfrac{d\theta_{i_M}}{ds} = M_W - M_{spr}  \\
    \label{eq:pure_mom_eq}
    \vspace{-4mm}
 \end{equation}   
 
  \begin{equation}
    \text{ and } E\cdot I\dfrac{d\theta_{i_F}}{ds} = F\cdot (L-\delta_x-X) 
\label{eq:pure_force_eq}      
  \end{equation}  
Where $\theta_i$ is the deflection of a bar in the ith layer, $\theta_{i_M}$ is the deflection of the bar under a pure moment and $\phi_{i_F}$ is the deflection of the under pure force. Equation \ref{eq:pure_force_eq} is solved following the procedure presented in Belendéz work \cite{Belen2002} and Equation \ref{eq:pure_mom_eq} is a linear equation that can be solved in a straightforward manner. The merged solution of both sub-problems is shown in Equation \ref{eq:sing_bar_sol}.

 \begin{align}
 \begin{split}
     M_{B_i} & = E\cdot I(\dfrac{\Gamma(\theta_{i_{F_0}})}{L_i} \cdot \sqrt{sin(\theta_0)} + \dfrac{\theta_{i_{M_0}}}{L_i}) \\
     &\text{where } \Gamma(\theta_{i_{F_0}})= \int_0^{\theta_{i_{F_0}}}\dfrac{d\theta}{sin(\theta_{i_0})-sin(\theta)} 
 \end{split}
\label{eq:sing_bar_sol}      
  \end{align}
 
  Where $\theta_{i_{F_0}}$ and $\theta_{i_{M_0}}$ are the force and moment contribution to the deflection $\theta_0$ at the tip of the bar and $L_i$ is the bent portion of the bar, which is indicated in Fig. \ref{fig:act_lengths}. Both deflection angles are not independent given that the moment $M_W$ occurs by the eccentricity of F about the upper attachment, relating the moment factor of both sub-equations \ref{eq:pure_mom_eq} and \ref{eq:pure_force_eq}. This relationship constrains both angles according to the constrain Equation \ref{eq:thts_const}:
  
   \begin{equation}
     \dfrac{I \cdot E}{2}\cdot (\theta_{i_{M_0}}-\Gamma(\theta_{i_{F_0}})^2 \cdot \dfrac{ecc}{2\cdot L_i})+\dfrac{M(spring)}{7}=0
\label{eq:thts_const}      
  \end{equation}

Finally, by replacing the bar moment and spring Equation \ref{eq:sing_bar_sol} and \ref{eq:spring_eq} in the moment balance Equation \ref{eq:mom_bal}, an equation that links the elastic mechanism's bending angle and generated moment are obtained.

\subsection{Experimental prototype of the multi-dof elastic mechanism}
Based on the previous model, a real-scale prototype was fabricated to evaluate the function of the multi-dof elastic mechanism.
For this application, 1.5 mm diameter bars made of carbon fiber with a bendable length of 80 mm were chosen. The bars were provided by Uxcell and the elastic modulus was determined experimentally as $80\pm10$ GPa.  
The reason for choosing carbon fiber is because of its low hysteresis among the conventional alternatives, according to a previous study \cite{chang2020}. In this context, hysteresis refers to the loss of energy during the loading and unloading of an elastic element \cite{chang2020}. The bars were joined to the attachment using quick-setting steel reinforced Epoxy from the brand J-B for being a simple and compact type of union.   

In each bar, a pair of stainless steel compression springs of 0.3 mm wire diameter, 20 mm height, and 5 mm of coil diameter were chosen. The spring constant was determined experimentally as $1.81\pm 0.01$ N/mm. The reason for using two springs instead of one was to prevent reaching the adhesion length when the bars are bent at the maximum angle and the spring has a high preload.

\begin{figure}
     \centering
     \begin{subfigure}[b]{0.2\textwidth}
         \centering
         \includegraphics[width=\textwidth]{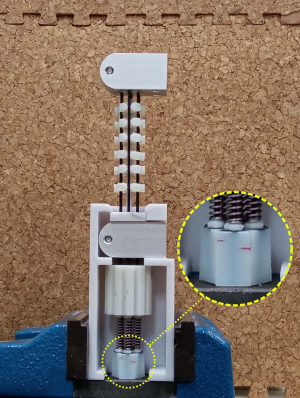}
         \caption{Extended condition}
         \label{fig:ext_proto}
     \end{subfigure}
     \hfill
     \begin{subfigure}[b]{0.2\textwidth}
         \centering
         \includegraphics[width=\textwidth]{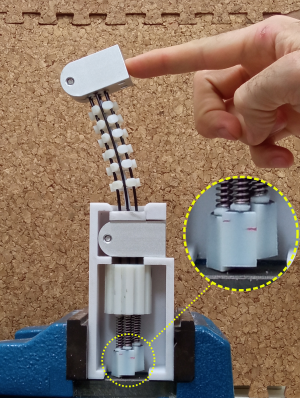}
         \caption{Bent condition}
         \label{fig:bent_proto}
     \end{subfigure}
        \caption{Real scale prototype. The detail in each figure shows the spring compression when the bars are bent.}
        \label{fig:real_proto}
        \vspace{-4mm}
\end{figure}

Most of the prototype structure was manufactured using 3d printed parts. The case and supportive components were printed in a Fortus 350, Stratasys, using PolyCarbonate filament. The mobile parts were fabricated in an Object Connex 250, Stratasys, employing the material "Vero White", which produces parts with smother surfaces to reduce friction in moving parts of the mechanism. Additionally, all the moving parts were lubricated with a silicon-based grease from Super Lube, and 2 mm diameter PTFE tubes were placed between the rods and their guides.
Fig. \ref{fig:real_proto} shows the finished prototype exhibiting the spring compression effect when the bars are bent.

\section{Experiments}
\subsection{Elastic mechanism bending test}
This experiment evaluates the elastic mechanism's stiffness with a bending test to compare the obtained stiffness with the theoretical estimation.
\subsubsection{Experimental setup}
\begin{figure}
     \centering
     \begin{subfigure}[b]{0.22\textwidth}
         \centering
         \includegraphics[width=\textwidth]{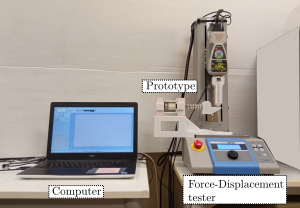}
         \caption{Main components of the experiment}
         \label{fig:ex_set}
     \end{subfigure}
     \hfill
     \begin{subfigure}[b]{0.22\textwidth}
         \centering
         \includegraphics[width=\textwidth]{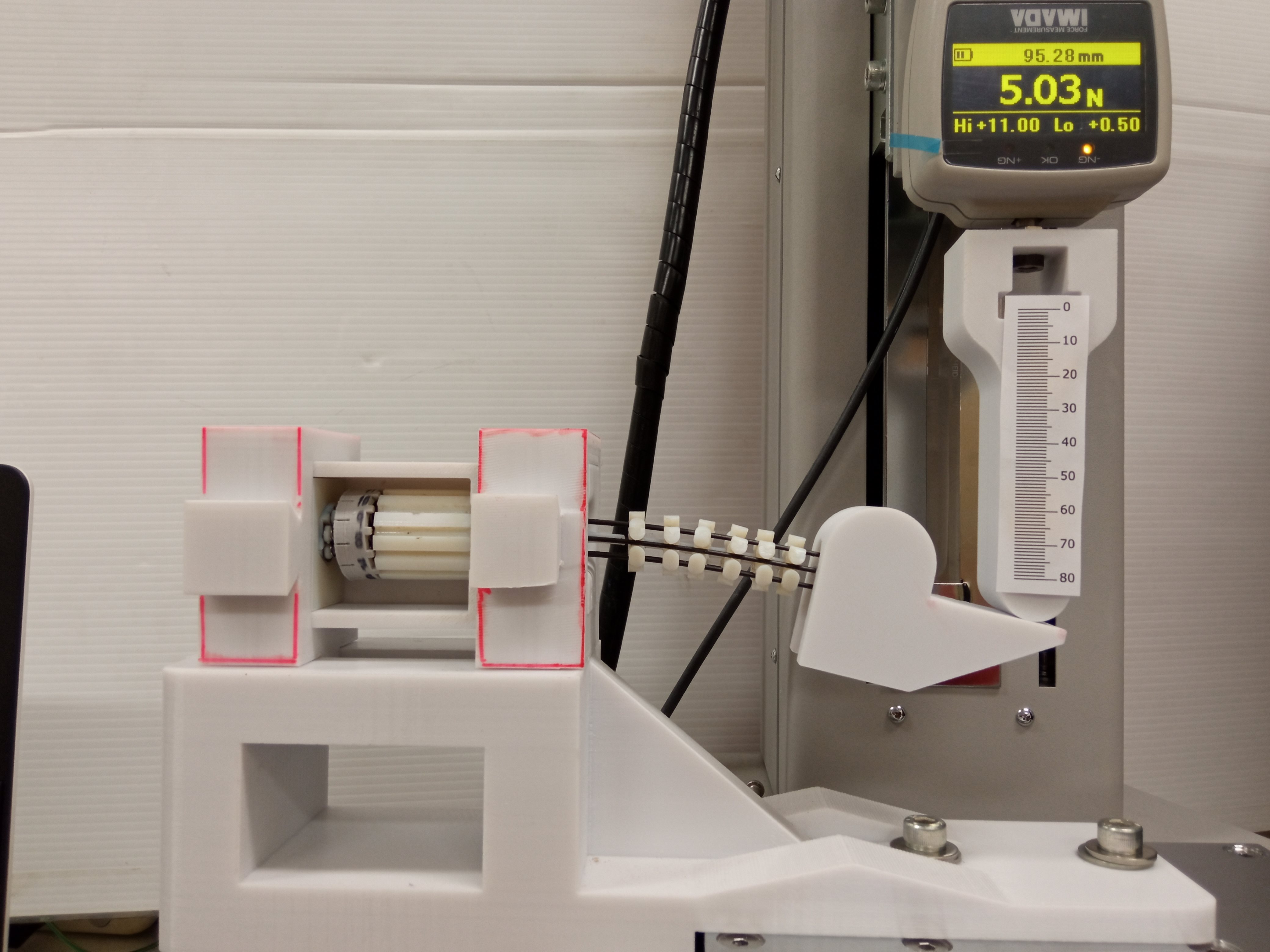}
         \caption{Bent prototype during test}
         \label{fig:bent_proto_tst}
     \end{subfigure}
        \caption{Experimental setup for the bending test.}
        \label{fig:exp_setup}
      \vspace{-4mm}
\end{figure}

 For performing the bending test, the prototype is mounted in a standard motorized test stand from IMADA with a force gauge and a displacement sensor (IMADA MX2-2500N-FA, ZTA-50N). The test stand was used to apply controlled loads to the upper attachment of the elastic mechanism. From the measurements, the elastic mechanism's assistive moment is calculated $H_{W_X}\cdot L_h$ as defined in the modeling section and is compared with the model estimation. The lever of the applied force and other geometrical parameters are measured by placing markers, recording the experiment, and analyzing the trajectories in Tracker, which is a software for performing motion capture analysis. The experimental setup along with the relevant geometrical parameters are shown in Fig. \ref{fig:exp_setup}. 

\subsubsection{Protocol}
The protocol consists in applying a loading sequence of three cycles. The elastic mechanism was bent for 9 mm of displacement corresponding to 16$^{\circ}$ of deflection approximately. This sequence was repeated for different stiffness configurations by adjusting the mobile plate that compresses the spring. The springs' preload was increased by 2.5 mm between sequences starting from the lowest stiffness mode.

\subsection{User experiment}
The effects of the elastic mechanism on the neck motion and muscle activity of healthy users were studied by mounting the experimental prototype on a standard neck brace.  

The device used for testing consists of a commercial rigid halo neck brace (Brand: Hengshui Jingkang, model N211, Instrument classification: Class I) which incorporates the elastic mechanism prototype. A sliding mechanism was implemented to connect the head attachment to the mechanism's upper part. It consists of a set of carbon steel rods that slide through copper thrusts when the user bends their head. A universal joint was added between the sliding mechanism and the upper attachment of the elastic mechanism to isolate moments generated by the device from the user's neck.

\begin{figure}[t]
    \begin{center}
    \includegraphics[width=0.9\linewidth]{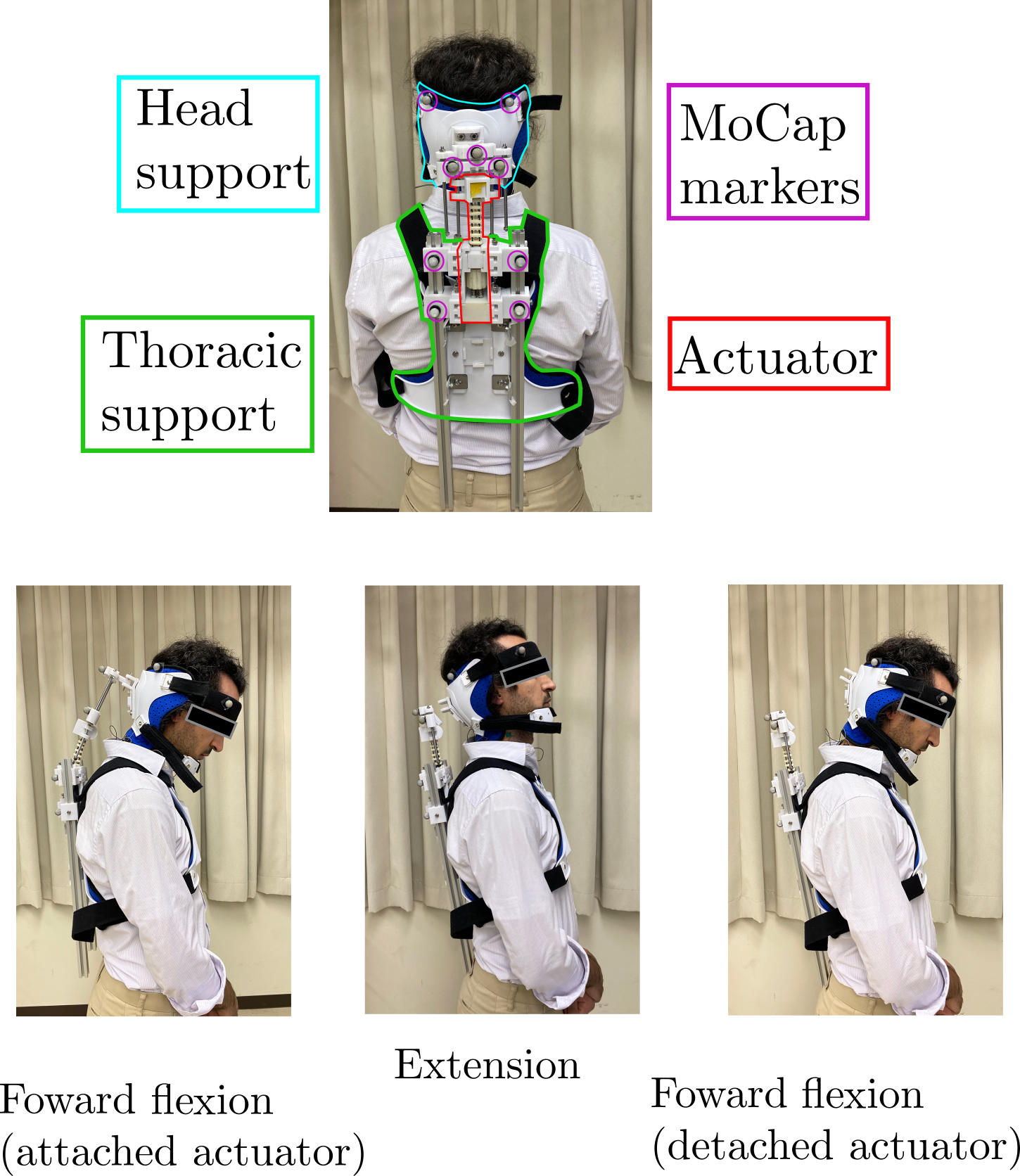} 
    \caption{User experiment parts and conditions when the user bends their head foward.}
    \label{fig:user_exp_conds}
    \end{center}     
    \vspace{-4mm}
\end{figure}

\begin{figure}[t]
    \begin{center}
    \includegraphics[width=0.7\linewidth]{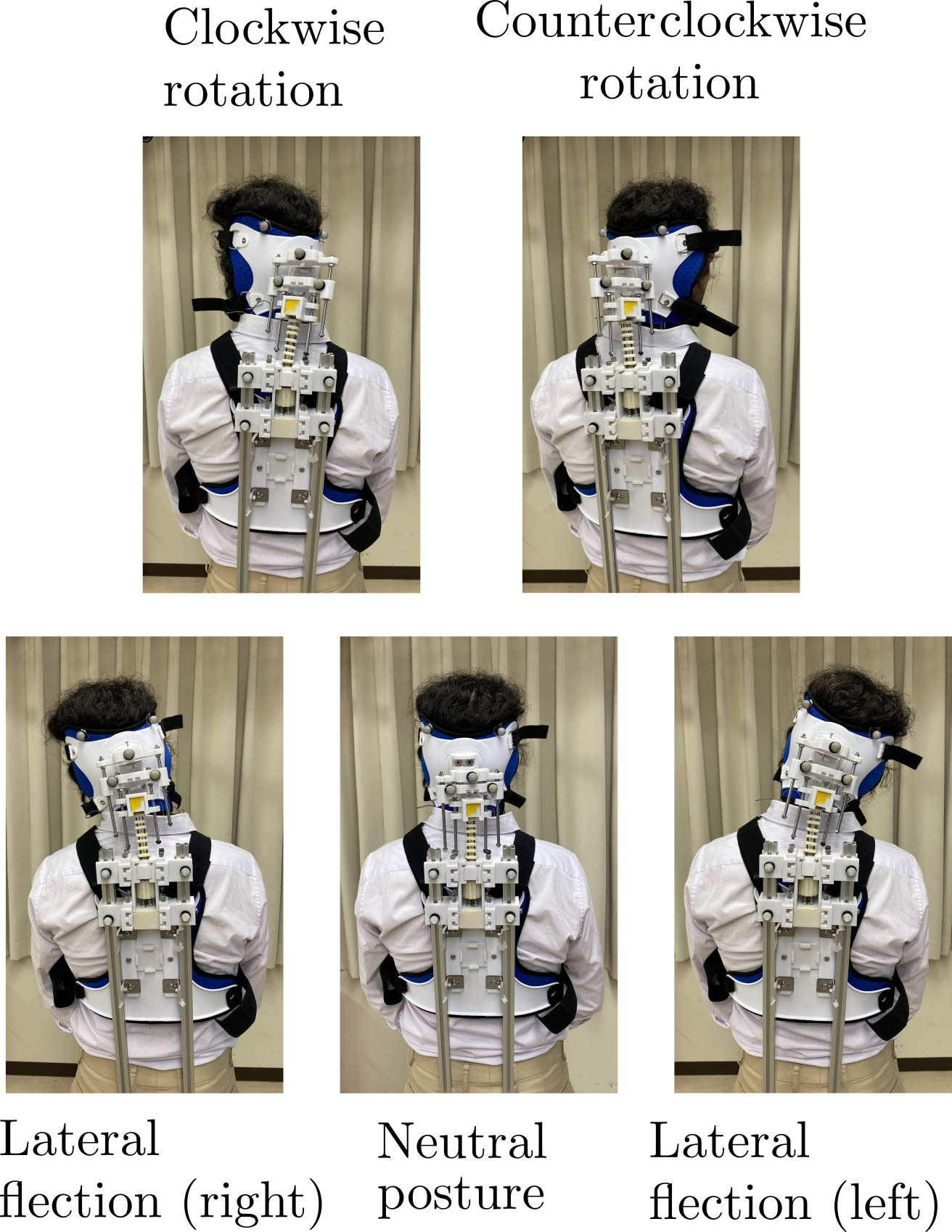} 
    \caption{Neck motions performed by the participant during the user experiment.}
    \label{fig:user_motions}
    \end{center}      
    \vspace{-4mm}
\end{figure}

We asked a healthy person with no history of neurological or musculoskeletal disorders to wear the exoskeleton and perform three times each of the movements shown in Figures \ref{fig:user_exp_conds} and \ref{fig:user_motions}: lateral neck flexions and extensions, sagittal neck flexions and extensions, and neck rotations. The participant wore the reflective markers indicated in Figure \ref{fig:user_exp_conds} to record the movements using a VICOM system. The participant was instructed to perform the task naturally without the exoskeleton, and reach the limit of the range of motion that did not cause strong discomfort with the exoskeleton. Each task's duration was instructed by the experimenter as 5 seconds for movement, and 5 seconds to hold the end position. The device and the experimental setup are presented in Fig. \ref{fig:exp_setup}.

%

%
\section{\uppercase{Results}}
\subsection{Bending test}
\begin{figure}[t]
    \begin{center}
    \includegraphics[width=0.9\linewidth]{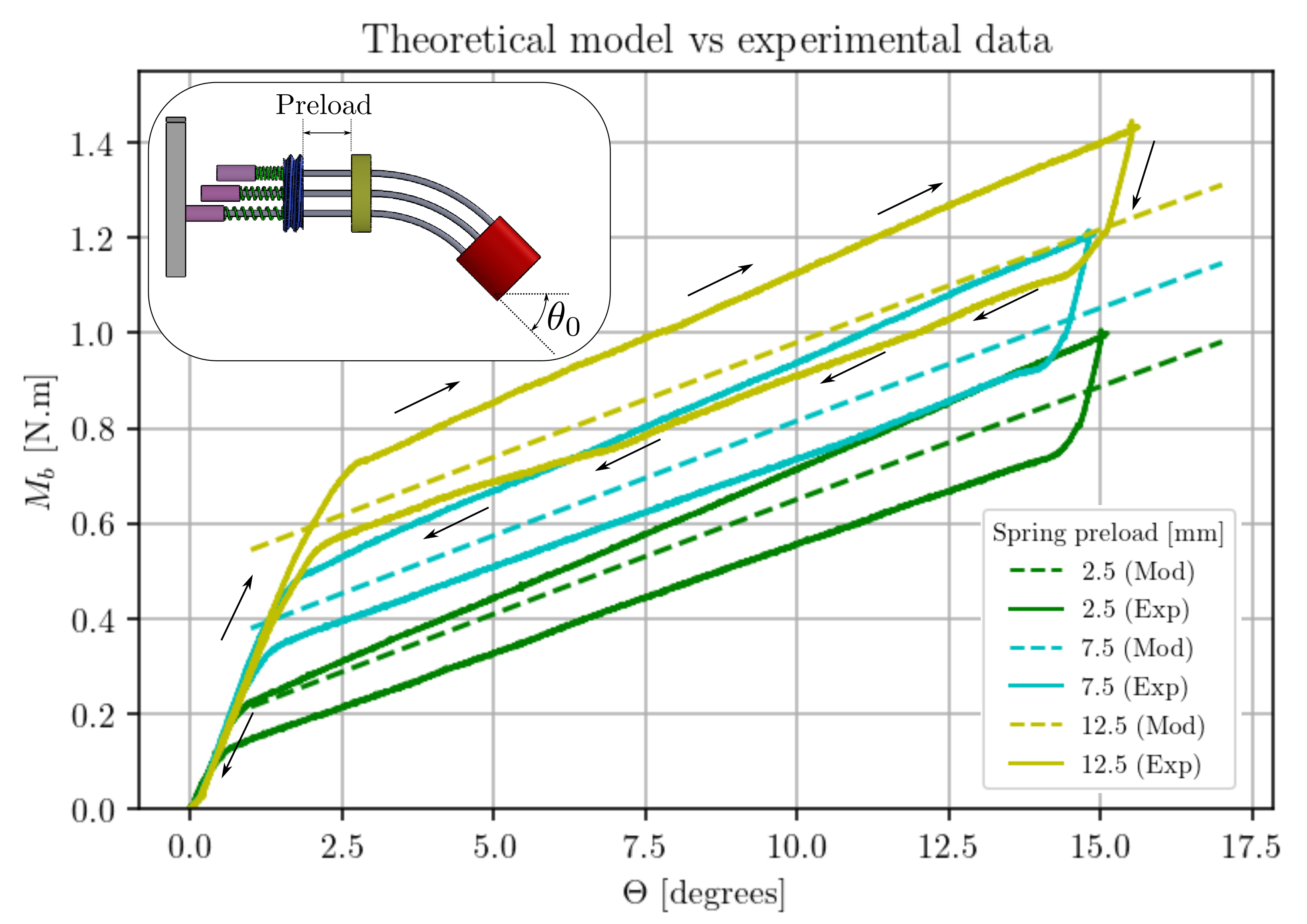} 
    \caption{The elastic mechanism's stiffness compared against the model prediction for three springs preload conditions. Each case shows a hysteresis cycle and three well-differentiated elastic regions during a flexion-extension cycle. The model presents a linear moment-deflection response situated inside of the experiment hysteresis cycle.}
    \label{fig:mech_exp_res}
    \end{center}    
    \vspace{-6mm}
\end{figure}

The theoretical and experimental bending stiffness of the elastic mechanism is contrasted in Fig. \ref{fig:mech_exp_res} for different spring responses. For better readability of the graph, only three of the six preload conditions are presented, given that the experimental data overlapped between consecutive conditions. The results show that for the three cases, the elastic mechanism presented three differentiated elastic regions.\\
When the elastic mechanism was initially extended, it exhibited a pure quasi-linear elastic behavior for low deflection, along the range which increases with the springs' preload. After surpassing the deformation threshold, which increases with the preload, the elastic mechanism behavior becomes linear with a lower rigidity than in the first stage. When unloading the mechanism, it initially presents a similar elastic behavior to the first stage and then a transition occurs where the stiffness becomes similar to the second stage accompanied by hysteresis behavior. When the mechanism is nearly fully extended, it behaves as in the first stage again.\\
On the other side, the model does not consider the mechanical loading history of the system and presents purely linear elastic behavior which is contained within the corresponding hysteresis cycle.

\subsection{User experiment}
\begin{figure}[t]
    \begin{center}
    \includegraphics[width=0.85\linewidth]{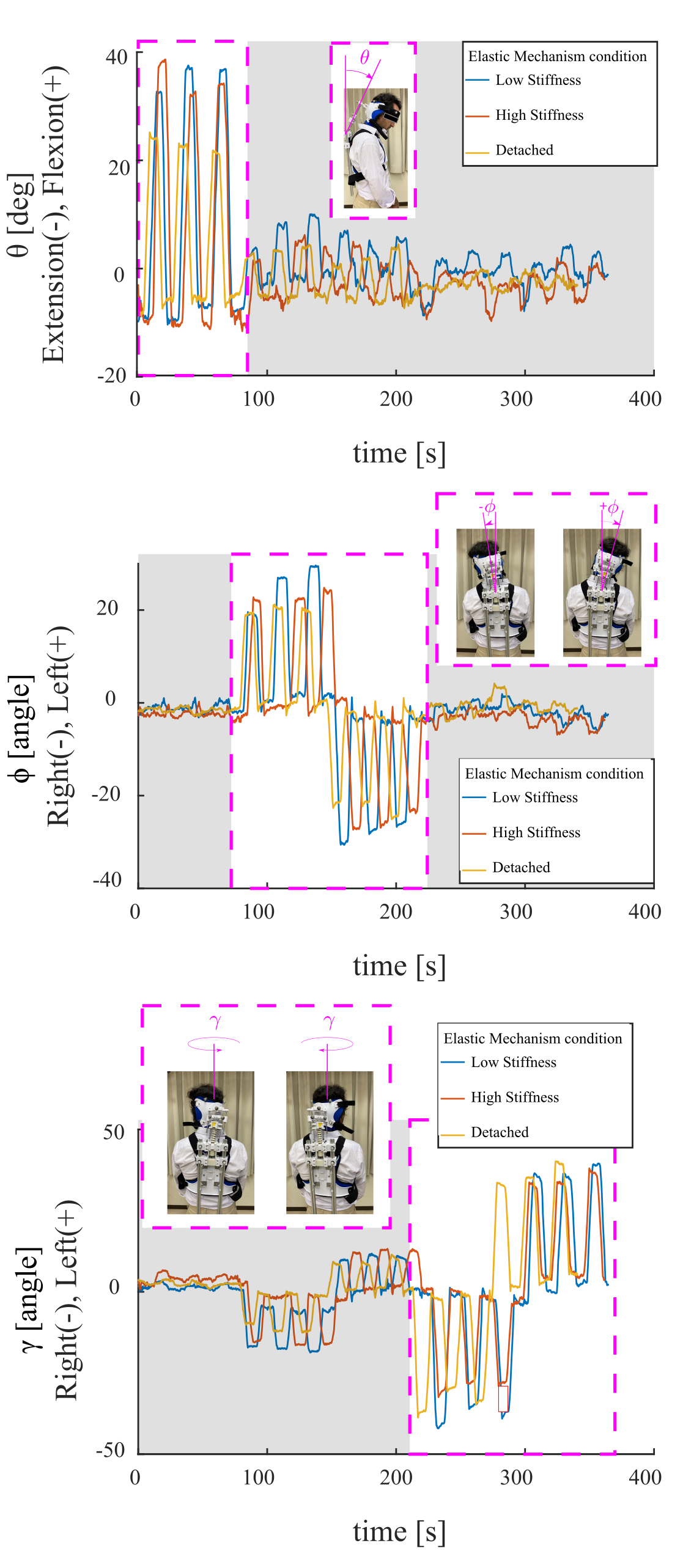} 
    \caption{Head range of motion for each condition during the user experiment. The magenta rectangle indicates the head motion associated with the given angle.}
    \label{fig:angle}
    \end{center}
    \vspace{-7mm}
\end{figure}

The motion sequence that the participant performed during the experiment for each condition is presented in Fig. \ref{fig:angle}. The figure shows the head orientation decomposed in three angles: flexion and extension in the sagittal plane, flexion and extension in the coronal plane, and the rotation about the head's vertical axis. Each of the mentioned angles is the main component of the head rotation during forward rotation, lateral bending, and axial rotation, respectively. The occurrence of each motion is enclosed with the magenta rectangle in each graph.
The participant was not able to perform neck extensions in the sagittal plane due to a collision between the sliding mechanism of the head attachment with the frame of the exoskeleton. Therefore, Fig. \ref{fig:angle}(a) only shows neck flexion movements. Other than neck extensions in the sagittal plane, the graphs show that the participant was able to achieve a similar range of motion with the exoskeleton and elastic mechanism to his normal neck articulations without major discomfort as self-reported by the participant. The graph shows that the "High stiffness" configuration of the elastic mechanism reduced the range of motion in corresponding neck articulations compared to the "Medium stiffness" configuration. This is expected since a higher stiffness will further restrict the flexion and rotation of the user's neck. Table \ref{tab:exp_rom} compares the range of motion among the three conditions.  
This experiment verifies the multi-dof articulation of the elastic mechanism when used in a neck support exoskeleton, and its limited impact on the neck range of motion.

\begin{table}[hbtp]
\begin{tabular}{|l|l|l|l|l}
\cline{1-4}
\textbf{Condition} & Flex. / Exten. & Lat. Bending & Ax. Rotation  &  \\ \cline{1-4}
Detached                  & $-9/22.2 ^\circ$    & $-22/22^\circ$  & $-41/36^\circ$  &  \\ \cline{1-4}
Low Stiff.             & $-7/38 ^\circ$      & $-23/28^\circ$  & $ -42/36^\circ$ &  \\ \cline{1-4}
High Stiff.            & $-8/32^\circ$       & $-25/22^\circ$  & $-34/33^\circ$  &  \\ \cline{1-4}
\end{tabular}
\caption{Maximum head inclination reached by the participant for each condition}
\label{tab:exp_rom}
\vspace{-3mm}
\end{table}


\section{Discussion}
The bending test serves as a proof of concept for regulating the stiffness of the elastic mechanism by the preloading of the compression springs. Also, it provides evidence that the model prediction is coherent with the real behavior of the elastic mechanism under similar preloading conditions. The validated model also permits the selection of a combination of materials and geometries of the compliant components to meet the assistive force criteria. 

The current prototype can provide an estimated force that is 38$\%$ of the force needed to fully compensate for the head weight and 12$\%$ of this force can be adjusted with the springs by using a stiffer model. For example, a spring with 0.9 mm wire diameter, 40 mm height, and 8 mm coil diameter will quadruplicate the previous value for a total compensation force of 66$\%$. These estimations are made assuming that the user's head weight is 5 kg and measuring $L_h$ module and $H_{W_x}$ direction according to Fig. \ref{fig:act_FBD} from measurements of the user wearing the device. The head orientation that requires the greatest assistive force was determined from a motion capture analysis of a healthy person doing the motions described in the user experiment protocol. As the assist value depends strongly on the patient's condition, the components will be defined based on the target user group in future work.\\

The difference between the model and the experiment is attributed mainly to friction on the moving parts and the deformation of the components apart from the spring and bars, which are not considered in the theoretical formulation. 
The three elastic regions mentioned in the result section are likely to be caused by these two factors.\\
In the first stage, the springs do not compress but the other components such as the attachments and unions deform generating the hysteresis behavior. After the transition point, the spring load surpasses the preload value which causes the compression of the springs. At this stage, the hysteresis is likely because of the friction between the bars and the PTFE tubes. This phenomenon along with the ideal rigid component deformation may explain the different slopes between the loading and unloading phases.\\
During loading, spring compression and component deformation happen together, but not during unloading. When unloading, component deformation starts recovering immediately, but the spring extension begins only after surpassing compression force and friction. As a result, the elastic mechanism extends with a short rigid recovery from the components followed by a longer and softer stage linked to springs with lower stiffness than loading.  

The user experiment provided evidence that the actuator does not impose severe restrictions on head mobility given that the range of motion does not differ significantly between the three conditions. 
Although the sagittal neck extension was restricted due to a design oversize, it is expected that the results will not differ significantly from the degrees of freedom obtained in the experiment. In this experiment, we did not study the neck muscle activation or the effects of using the device for prolonged periods. These investigations will be conducted in future works with an increased number of participants and a revised design of the exoskeleton, before any evaluation on end users. 

\section{Conclusion}
This work presented the design, development, modeling, and testing of a passive compliant neck exoskeleton based on an elastic mechanism with variable stiffness capabilities. A physical model considering the compliant element's mechanical properties was made to emulate the elastic mechanism behavior and was experimentally validated. The model prediction and experimental validation suggest that the elastic mechanism has the potential to meet the biomechanical requirements for clinical application.   

\addtolength{\textheight}{0cm}   

\section*{ACKNOWLEDGMENT}
This work was supported by the Ministry of Education, Culture, Sports, Science and Technology (MEXT) of Japan.

\bibliographystyle{IEEEtran}
\bibliography{IEEEabrv,MyCitations}
\end{document}